\documentclass[pdflatex,sn-mathphys-ay]{sn-jnl}

\usepackage{graphicx}
\usepackage{multirow}
\usepackage{amsmath,amssymb,amsfonts}
\usepackage{amsthm}
\usepackage{mathrsfs}
\usepackage[title]{appendix}
\usepackage{xcolor}
\usepackage{textcomp}
\usepackage{manyfoot}
\usepackage{booktabs}
\usepackage[normalem]{ulem}
\usepackage[capitalise]{cleveref}

\theoremstyle{thmstyleone}

\theoremstyle{thmstyletwo}

\theoremstyle{thmstylethree}

\newcommand{\textbook}{\emph{textbook}}
\newcommand{\dset}{D}
\newcommand{\divscore}{\mathcal{V}}
\newcommand{\frep}{\xi}

\begin{document}


\title[]{




Exemplar-based objective classification of gust-induced loads across multiple flight conditions

}
\author*[1]{\fnm{Paolo} \sur{Olivucci}}\email{p.olivucci@tu-braunschweig.de}

\author[1]{\fnm{Kowshik} \sur{Srivatsan}}\email{k.srivatsan@tu-braunschweig.de}

\author*[1]{\fnm{David E.} \sur{Rival}}\email{david.rival@tu-braunschweig.de}

\affil*[1]{%
  \orgname{Institute of Fluid Mechanics, TU Braunschweig}, \orgaddress{%
    \street{Hermann-Blenk Str. 37}, \city{Braunschweig}, \postcode{38108}, \country{Germany}}}

\abstract{
Is it possible to find an objective classification criterion that organizes the complexity of
gust-induced loads across many flight conditions? 
And one that remains as interpretable as a labelling based on coarse parameters, such as the flight
attitude? 
Our approach encodes a large number of experimental observations through a machine-learned
representation and applies a summarization procedure to select a minimal subset of highly
significant exemplars. 
The exemplars provide a similarity-based objective classification criterion of all the observations,
they can be more conveniently inspected by experts and can become subject of more refined
experiments. 
We demonstrate the approach on a database of 3480 pressure-load measurements induced by random gusts
on a flying-wing model across six flight attitudes. 
We find nine fundamental response types that recur across multiple attitudes; 
analysis of a type's transient response enables physical intuition into the underlying fluid
mechanics. 

}

\maketitle


\section{Introduction}
\label{sec:intro}


Small uncrewed aerial vehicles (UAVs) are increasingly deployed in low-altitude environments where
atmospheric turbulence is spatio-temporally complex. In this scenario, the length and velocity
scales of unsteady disturbances overlap directly with the vehicle dimensions, which makes
gust-induced load excursions a critical threat to flight stability and structural integrity
\citep{Jones2022, Floreano2015}. The gust-wing interaction parameter space is very high-dimensional,
spanning manifold combinations atmospheric conditions, terrain features and vehicle attitude.
Further, the resulting aerodynamic response is typically strongly nonlinear, which contributes to
making interactions of gusts with small aircraft poorly understood \citep{Jones2021, Marzanek2019}. 

Traditional gust models, from early deterministic velocity profiles to modern
statistical descriptions of atmospheric turbulence, capture only a limited slice of this complexity
\citep{Fuller1995, Moorhouse1982}.
The classical linear framework of indicial response theory provides a closed-form description of
unsteady lift under small-amplitude disturbances \citep{Leishman1996}.
However, UAVs in the lower atmospheric boundary-layer largely operate outside the linear regime. 
For instance, the aerodynamics of non-slender delta wings is dominated by leading-edge vortex (LEV)
development and strongly nonlinear separation phenomena~\citep{Gursul2005, Marzanek2019}, under which
linear superposition fails. 
Identifying an organizing criterion for the complexity of gust-load responses in this regime,
therefore, calls for an approach grounded in empirical data before analytical theory. 

Recent work in data-driven fluid mechanics show a promising trend in this sense.
Sparse pressure sensing strategies have been developed for real-time load estimation during gust
encounters~\citep{Chen2023, Iacobello2022}, and deep autoencoders have compressed complex vortical
interactions onto low-dimensional manifolds with physically interpretable
coordinates~\citep{Fukami2023}. 
Cluster-based approaches have also shown promise for flow-state estimation from sparse experimental
measurements~\citep{Kaiser2024}. 
Results from the machine learning literature have demonstrated that data quality often matters more
than quantity; predictive models trained on curated, high-value training sets can outperform those
trained on orders of magnitude more unfiltered data~\citep{Sener2018,Gunasekar2023}. 
This motivates a sparse-in-the-data approach: a selected subset of the most informative gust
encounters may suffice to capture the essential physics and enable understanding and further
analysis.  
In this respect we build on previous work\footnote{Currently under journal review.}
\citep{Olivucci2026}, that demonstrated the feasibility of finding a minimal subset, dubbed 
\textbook{}, of a database of gust-wing encounters at a single angle of incidence with
minimal consequences on load prediction accuracy. 

\subsection{Objectives and approach}
The central motivation of this study is the question on what constitutes an objective criterion for
classifying gust-induced loads in a complex flight envelope. 
Specifically, we ask whether an interpretable classification criterion such as labeling loads based
on the vehicle attitude is adequate, or whether an criterion can be identified from available
observations without losing ease of being understood by human experts.  

\begin{figure}[h]
\centering
\includegraphics[width=0.59\linewidth]{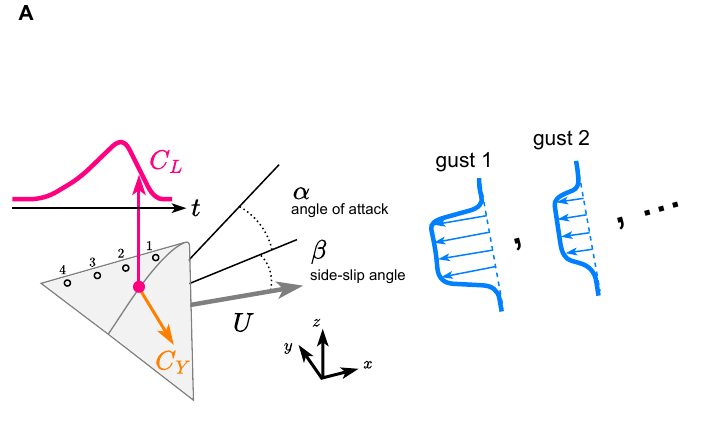}
\hfill
\includegraphics[width=0.39\linewidth]{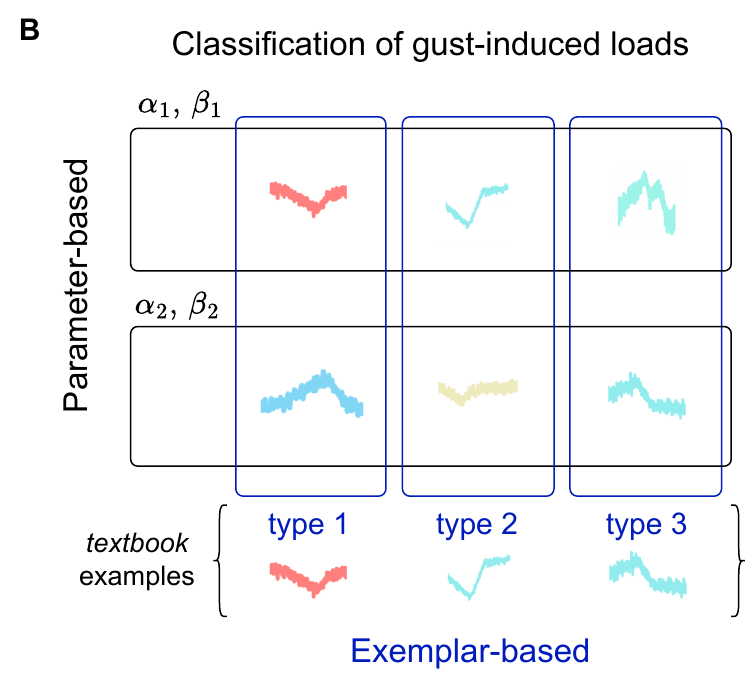}
\caption{\textbf{Gust-induced loads.}
(a)~A flying wing encounters a series of gusts. 
As the wing's attitude varies, each encounter can happen at a different orientation. 
(b)~ Classification of gust-induced load response according to two different criteria:
parametric and exemplar-based. 
}
\label{fig:gust_loads_attitudes}
\end{figure}

The problem setting is illustrated in \cref{fig:gust_loads_attitudes}a. 
The case we consider is that of a non-slender delta wing encountering a number of longitudinal gusts
at several incidence configurations. 
The choice of a non-slender delta wing is well suited to investigating complex loads, as its
aerodynamics is dominated by the development of unsteady nonlinear phenomena such as leading-edge
vortices (LEVs). 

Our proposed approach is conceptualized in \cref{fig:gust_loads_attitudes}b and contrasts two
classification criteria; the first is based on human-defined parameters, the other is based on
labeling loads according to a set of representative exemplars, called \textbook{} examples. 

The flight attitude is an important determinant of the aerodynamic loads acting on a wing under
given atmospheric conditions and therefore constitutes a natural candidate as a tentative
classification criterion of the gust-load responses. 
It is sensible to ask what extent the flight attitude explains the diversity of the load response
and then interrogate the data to find an objective criterion. 
Successfully finding such an organizing criterion could serve as the basis for understanding the
response phenomenology and for investigating its roots in the fluid dynamics of gust-wing
interaction. 

 
In  practice, our contribution consists of searching for this criterion on a sizable amount of
experimental gust-load measurements across several flight attitudes (3479 encounters across 6
attitudes) and extracting a small number of significant exemplars in an objective manner via a
learned predictive model. 
These \textbook{} examples are taken as the canonical gust-induced load types and used to classify
all observations. 
This exemplar-based classification criterion is then compared to the one based on the incidence
angles and utilized to guide the search for the fluid mechanical origins of different gust-load
types. 

The article is laid out as follows. 
The experimental facility and the aerodynamic load database are described in \cref{sec:experiment}
and \cref{sec:database}.  
The data summarization approach is introduced in \cref{sec:method} and its concrete realization in
\cref{sec:textbook_method}. 
The resulting \textbook{} examples are in \cref{sec:textbook_results} and applied to gust-load
classification in \cref{sec:patterns}. 
The fluid dynamical characterization of representative cases is considered in
\cref{sec:characterization}. 

\section{Experimental setup and database}
\label{sec:experiment}

\subsection{Random gust generator}
\label{sec:facility}

\begin{figure}[h]
\centering
\includegraphics[width=\linewidth]{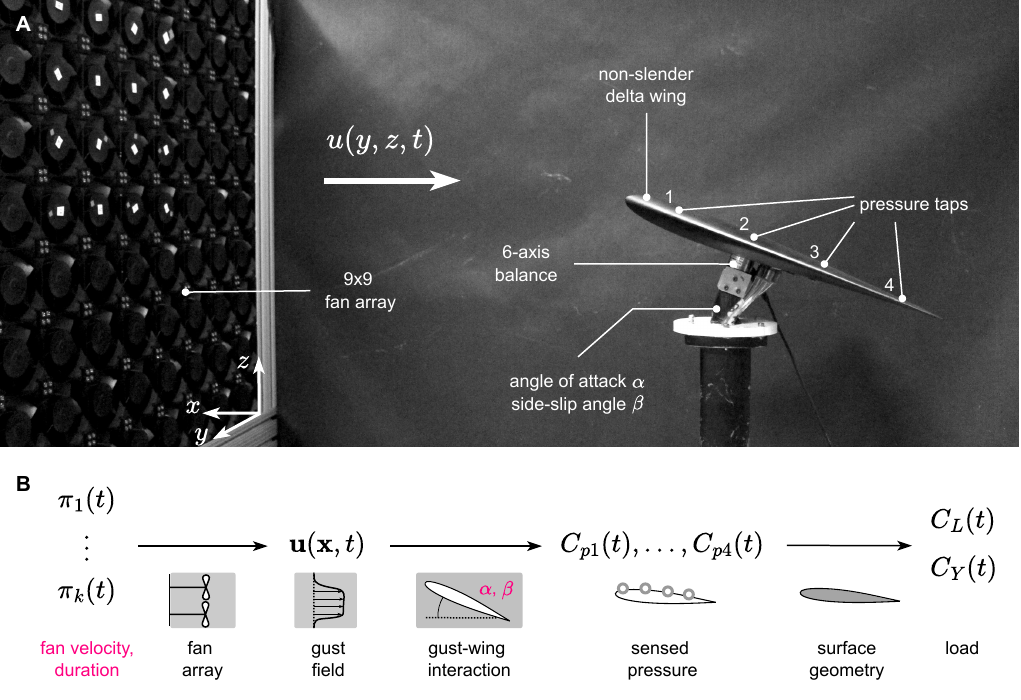}
\caption{\textbf{Random gust generator and experimental setup.}
(a)~The fan-array facility with the non-slender delta wing mounted on its sting, showing pressure
taps and 6-axis balance. 
(b)~Causal chain linking the experimental parameters (the forcing parameters $\pi_k(t)$ and the
incidence angles $\alpha, \beta$, highlighted in red) to the aerodynamic loads $C_L(t)$ and
$C_Y(t)$.  
}  
\label{fig:facility}
\end{figure}

The experimental facility, shown in \cref{fig:facility}a, consists of a 9 $\times$ 9 array of
computer-controlled dual tube-axial DC fans capable of generating unsteady axial inflow across a
broad range of conditions. 
The facility can be programmed to operate semi-autonomously, performing a high volume of trials
and enabling coverage of a large experimental space spanned by combinations of many forcing
parameters. 

The aerodynamic model is a non-slender delta wing with a NACA0012 cross-section and a mid-span of
chord $c = 30$cm, adapted from previous work \citep{Marzanek2019, Burelle2020}. 
Four pressure taps are positioned near the leading edge at locations determined by a dedicated
optimisation study \citep{Chen2023}. 
A six-component force balance mounted on the support sting records the time-resolved aerodynamic
loads. 
The sting additionally allows independent adjustment of the static angle of attack $\alpha$ and
sideslip angle $\beta$, enabling systematic variation of the wing attitude. 
The causal chain linking the experimental parameters to the measured aerodynamic response is
illustrated schematically in \cref{fig:facility}b.  

The fan array is driven by three randomized forcing parameters: the base fan velocity $\pi_1$, the
velocity increment $\pi_2$, and the forcing interval duration $\pi_3$. During each interval, the
fans accelerate or decelerate from the base velocity toward the target increment, before the cycle
repeats. The base velocity spans 100 discrete levels, keeping the chord-based Reynolds number in the
range $6 \cdot 10^{4} < \frac{c U_\infty}{\nu} < 3.5 \cdot 10^{5}$, where $U_{\infty}$ denotes the
mean freestream speed produced by a fan velocity setting away from the wing. 
The velocity increment is drawn from a uniform random distribution between 10\% and 130\% of
the base fan velocity, and the non-dimensional forcing duration is randomized with a minimum value
of $G/c=2$. 
This randomized operation ensures unbiased coverage of the gust parameter space across successive
trials. 
 
\subsection{Gust-event database}
\label{sec:database}







The experimental database spans six vehicle attitudes, combining two angles of attack $\alpha \in
\{20^\circ, 30^\circ \}$ with three sideslip angles $\beta \in \{0^\circ, 15^\circ, 30^\circ\}$,
ensuring markedly distinct flow states on the wing.
The attitude $\alpha = 30^\circ, \beta = 0^\circ$ is represented by 100 one-minute trials, one for
every fan speed level. 
All remaining attitudes are represented by 50 trials each, yielding a total of 350 one-minute trials 
across the full database; an overview is given in \cref{tab:attitudes}. 

The four pressure tap readings and the full six-component balance output are recorded continuously
throughout each trial, of which the lift coefficient $C_L(t)$ and the side-force coefficient
$C_Y(t)$ capture the two dominant load components under combined angle-of-attack and sideslip
conditions. 
Each trial is a continuous signal that contains the response signatures to many randomly occurring
successive gust encounters, which we call ``events''. 
A time-series segmentation procedure is therefore applied to each trial to isolate individual gust
events as the elementary units of data (also known as ``token'').
The automated segmentation procedure is detailed in \cref{sec:segmentation} and is based on the lift
coefficient history $C_L(t)$, which serves as the primary indicator of aerodynamic response.  

\begin{table}[h]
\caption{Overview of the six flight attitudes and the 3479 events in the gust-load database.}
\label{tab:attitudes}
\begin{tabular}{llllll}
\toprule
$\alpha$ (deg) & $\beta$ (deg) & Trials & Event count & Duration avg, std ($t^*$) & Amplitude avg,
std \\
\midrule
20 & 0  & 50  & 502 & $114.6 \pm 41.2$ & $1.97 \pm 0.49$ \\
20 & 15 & 50  & 530 & $112.3 \pm 37.1$ & $1.76 \pm 0.47$ \\
20 & 30 & 50  & 512 & $114.7 \pm 40.6$ & $1.56 \pm 0.51$ \\
30 & 0  & 100 & 973 & $119.5 \pm 45.5$ & $1.74 \pm 0.46$ \\
30 & 15 & 50  & 473 & $115.8 \pm 42.6$ & $2.39 \pm 0.56$ \\
30 & 30 & 50  & 489 & $109.7 \pm 34.7$ & $1.97 \pm 0.54$ \\
\botrule
\end{tabular}
\end{table}

\begin{figure}[h]
\centering
\includegraphics[width=\linewidth]{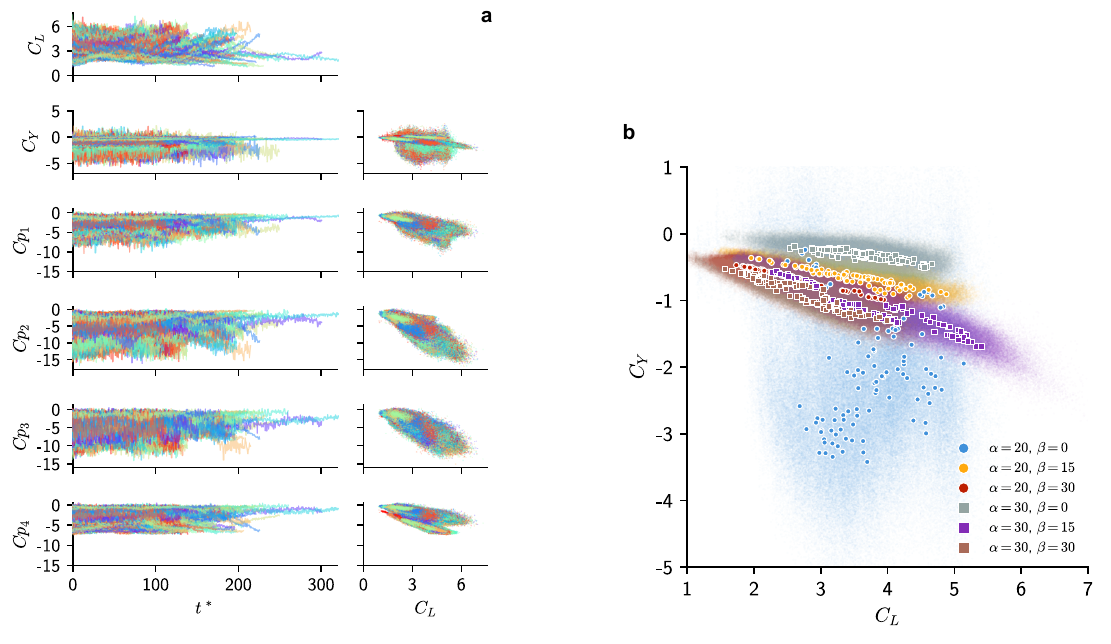}
\caption{\textbf{Gust-load event database.}
(a) event time-series (left-hand column) and their projection on each coordinate plane of the
input-response space (right-hand column).
Time is expressed in convective units $t^* = t U_\infty / c$. 
(b)~Distribution of the 2783 training gust events on the $C_L$--$C_Y$ response plane, colored by
flight attitude. 
Large symbols mark the centroids of each event and symbol shape distinguishes the two values of
$\alpha$. 
}
\label{fig:database}
\end{figure}

The final database comprises 3479 gust events; some descriptive statistics are summarized in
\cref{tab:attitudes} and the events are visualized in \cref{fig:database}. 
The mean event duration is approximately 115 convective times $t^* = t U_\infty/c$, with a standard
deviation of approximately 41 convective times. 
The database is partitioned into a training set of 2783 events and a test set of 696 events
following an 80--20 split. 

The left column of \cref{fig:database}a shows the time series of all 2783 training events for each
channel including the four pressure readings $C_{p1}(t)$ through $C_{p4}(t)$ and the two load
outputs $C_{L}(t)$ and $C_{Y}(t)$ overlaid across all the attitudes. 
The projections of each channel against $C_L$ reveal that the sensed pressure and the aerodynamic
loads have a non-linear dependence, with a non-trivial data geometry which reflects the complexity
of the underlying gust-wing interactions. 

\cref{fig:database}b provides a scaled-up view of the projection of the data on the
$C_{L}-C_{Y}$ response plane and also labels events by the flight attitude they were recorded at. 
The six attitudes occupy partially overlapping regions of the response space.
Increasing the sideslip angle $\beta$ produces a systematic shift toward more negative side-force
values $C_Y$ across both angles of attack, while the higher angle of attack $\alpha = 30^\circ$
extends to larger lift values. 
The data geometry in the response space raises the question of whether the attitude is a fundamental 
organizing parameter of gust-induced load response or whether objective response patterns that
transcend attitude boundaries can be identified. 


\section{Data-driven response modelling and summarization}
\label{sec:method}

The size and complexity of the database supplied by the experimental facility raises the issue of
extracting the information contained therein in a form accessible to a human expert. 
Compressing a large experimental sample to a concise, maximally informative subset of examples
would bridge the gap between large-scale automated data collection and physical understanding.  
Borrowing from a previous study, we refer to such a subset as a \textbook{} \citep{Olivucci2026}.

\subsection{\textbook{} datasets}\label{sec:textbook_datasets}

A \textbook{} is understood as a small subset $\dset_\text{txt} \subset \dset$ of a large database
$\dset$ that is 
accurate, in that it guarantees high generalization accuracy on unseen data;
and parsimonious, in that it is as small as possible while remaining accurate.
Formally, the first property can be defined as follows. 
Given a predictive model trained on subsets $\dset_m \subset \dset$ of cardinality $|\dset_m| = m$
with $m \ll |\dset| = m_\infty$ and evaluated on a held-out test set, the \textbook{}
of size $m$ is the subset that minimizes: 

\begin{equation}
\dset_\text{txt}(m) = \underset{\dset_m \subset \dset,\, |\dset_m| = m}{\arg\min}\, \varepsilon(\dset_m)
\label{eq:textbook_ideal}
\end{equation}

where $\varepsilon(\dset_{m})$ denote the test error of the model trained on $\dset_m$.
\cref{eq:textbook_ideal} defines a combinatorial search problem over all
$\binom{|\dset|}{m}$ possible subsets, which is computationally intractable for any dataset of
practical size.  
Surrogate approaches are therefore needed.

The computer science literature (where $\dset_\text{txt}$ are usually called \emph{core-sets})
identifies two tractable ways of constructing approximations to $\dset_\text{txt}$: one is based on
selecting individual samples of high value or ``difficulty'' for the model to learn
\citep{Paul2021,Koh17}; the other is to aim at good coverage of the sample diversity through
clustering-like procedures \citep{Sener2018}. 
Here we take the second path, which is empirically known to be superior for smaller subsets sizes.



The coverage approach necessitates two main elements.
The first is a representation function $F: \dset \rightarrow \mathbb{R}^d$  (an embedding)
that maps each event in the database to a fixed-length descriptor vector $\xi$ of dimension $d$.
The second is a diversity score $\divscore_\frep(\dset_{m})$ that measures the internal diversity of
subset $\dset_m$ by comparing its elements in $\xi$-space. 
The surrogate \textbook{} of \cref{eq:textbook_ideal} is then found as the subset that maximizes 

\begin{equation}
\dset_\text{txt}(m) \approx \underset{\dset_m \subset \dset,\, |\dset_m| = m}{\arg\max}\, \divscore_\frep(\dset_m),
\label{eq:textbook_surrogate}
\end{equation}

rewarding coverage of the full dataset while penalizing redundancy among selected events.
Crucially, the information captured by \cref{eq:textbook_surrogate} reflects the choice of embedding
coordinates $\frep$ and diversity score $\divscore$. 
An ideal \textbook{} would therefore use a representation that reflects the underlying functional
relation (in the present case, the pressure-load response).



Finally, the choice of \textbook{} size $m$ is based on a trade-off with predictive accuracy, in analogy
with lossy compression. 
If $\varepsilon(\dset)$ is the baseline test error when training on the full
database, the optimal \textbook{} size $m^*$ is the smallest subset that achieves performance
$\varepsilon\!\left(\dset_\text{txt}(m)\right) \geq \delta\cdot\varepsilon(\dset)$ within a desired
tolerance $\delta\in(0,1)$ of the full-database limit.


The specific choice of predictive model, representation $\frep$, diversity score $\divscore$ and
tolerance $\delta$ are described in \cref{sec:textbook_method}.

\subsection{Data summarization and \textbook{} selection} 
\label{sec:textbook_method}

The aerodynamic load prediction model used to validate \textbook{} quality is a Multi-Layer
Perceptron (MLP) which takes the four instantaneous pressure readings $C_{p1}$ through $C_{p4}$ as
inputs and predicting $C_L$ and $C_Y$ simultaneously as outputs. 
The model has two hidden layers with 32 ReLU units each.
The MLP is trained using the combined mean squared error (MSE) loss on $C_L$ and $C_Y$ as the loss
metric $\varepsilon$.
First, training is performed on the entire database to determine the baseline accuracy, then it is
repeated from scratch for each of the subsets of increasing size (selected as explained below),
resulting in a different set of trained weights for each subset. 
Each model is evaluated on the same held-out test set using the same MSE loss metric. 

\begin{figure}[h]
\centering
\includegraphics[width=\linewidth]{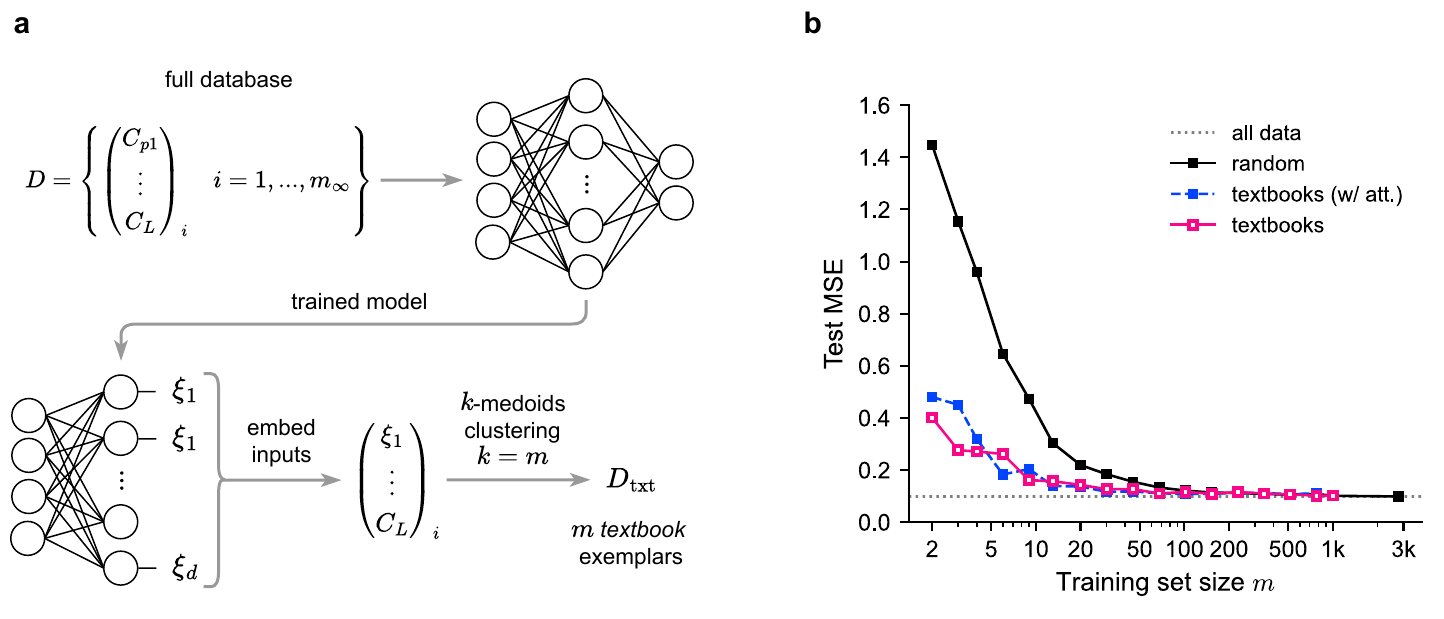}
\caption{\textbf{\textbook{} datasets.}
a) Data subset selection procedure. 
Symbols are as defined in \cref{sec:textbook_datasets} and the procedure explained in
\cref{sec:textbook_method}. 
b) Learning curves. 
Test error of models trained on increasingly larger subsets of the full dataset, selected according
to several criteria: 
randomly (black solid), 
\textbook{} subsets selected without attitude labels (red empty), 
and including attitude labels (blue solid).
The dotted line marks the baseline accuracy achieved by the full database. 
}
\label{fig:learning_curves}
\end{figure}

A schematic view of the \textbook{}-finding procedure is in \cref{fig:learning_curves}a.
As the embedding coordinates $\xi$ we take the output of the trained MLP truncated before its final
(linear) layer. 
This constitutes a set of learned coordinates that embed the network inputs into a linear manifold
of dimension equal to the layer width ($d=32$ in the present case). 
L2 distances between data point embeddings i.e. $(\xi, C_L, C_Y)$ are thus a natural candidate for
an objective, data-driven pairwise diversity criterion. 

The \textbook{} subsets are identified through a data summarization procedure based on $k$-medoids
clustering in embedding space (\cref{fig:learning_curves}a bottom row), in which the $m$ cluster
centers constitute the \textbook{} events. 
Formally, this amounts to solving \cref{eq:textbook_surrogate} with $\divscore(\dset_m)$ defined as
the sum of the average distances from cluster centers \citep{Tansel1983}. 
The pairwise similarity between events is the L2 distance between their barycenters in $\xi$-space. 

The performance of \textbook{}s selected according to this procedure is benchmarked against random
subsets of equivalent size, which serve as a reference for the expected accuracy of unguided subset
selection.
The resulting learning curves are discussed in \cref{sec:textbook_results}.

The optimal \textbook{} size $m^*$ is selected using a tolerance identified from the learning curve
as the point of diminishing returns, beyond which additional \textbook{} events yield negligible
improvement in test accuracy and is reported in \cref{sec:textbook_results}.  

\section{Results}
\label{sec:results}

\subsection{\textbook{} selection}
\label{sec:textbook_results}



The learning curves resulting from the selection procedure for random subsets, labeled
\textbook{}s, and unlabeled \textbook{}s are shown in \cref{fig:learning_curves}b. Across all
training set sizes, both \textbook{} conditions achieve substantially lower test error than random
subsets of equivalent size, confirming that the summarization procedure successfully identifies
high-information events. 
At the chosen \textbook{} size of $m^* = 9$, the \textbook{} achieves a test mean-squared error of
0.161, compared to 0.473 for a random subset of the same size which is a reduction of approximately
66\%, while the full-database baseline stands at 0.099.  
This corresponds to a compression ratio of $m^*/m_\infty \approx 0.3\%$, representing a reduction of
the full training set size by over two orders of magnitude, and about $15\%$ when compared to
random subsets of size $60$ that yield comparable test accuracy.

In order to assess to which extent the attitude provides a sensible classification criterion for
gust-load response, \textbook{} selection is repeated with the inclusion of attitude labels for each
event.  
In this setting, the subset selection algorithm will select that  diverse attitude make-up in terms of
the attitude.  
If the two subset selection setting yield \textbook{}s of comparable generalization accuracy, it
follows that attitude labeling is superfluous for the purpose of capturing the essential diversity
of the data. 

It is apparent in \cref{fig:learning_curves}b that the learning curves for attitude-labeled and
unlabeled \textbook{}s do not show diverging trends to a significant level. 
At $m^* = 9$ the unlabeled condition marginally outperforms the labeled one, while at $m = 13$ and
$m = 20$ the two conditions converge to within 5\% of each other. 
The absence of any significant effect from attitude labeling indicates that the summarization
procedure already incorporates the essential data diversity, including in the attitude.

\subsection{Cross-attitude response}
\label{sec:patterns}

\begin{figure}[h]
\centering
\includegraphics[width=\linewidth]{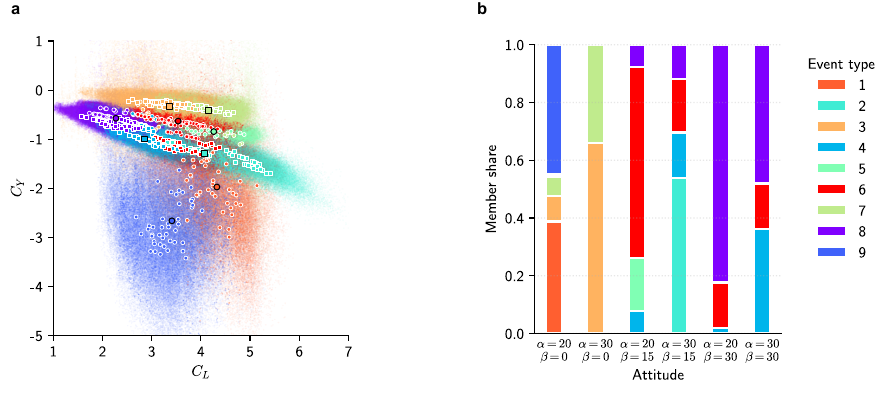}
\caption{\textbf{Objective classification of gust-load events.}
(a)~Distribution of all events in response space, colored by type as identified through clustering
around the \textbook{} events. Large symbols represent \textbook{} event centroids as in
\cref{fig:database}b. (b)~Cluster membership share per attitude, showing which response types are
attitude-specific. }
\label{fig:clusters}
\end{figure}

Each of the $m^*=9$ \textbook{} events serves as a representative exemplar of a response type within
the full gust-load database, providing the objective labeling criterion theorized in
\cref{fig:gust_loads_attitudes}b. All the 3479 gust-load events are assigned to their nearest
neighboring \textbook{} event through the same similarity criterion used during selection. 
The resulting clustering is visualized in \cref{fig:clusters}a, which shows the same $C_L-C_Y$
response space as \cref{fig:database}b but now with each event colored by assigned response type
rather than flight attitude.


The nine clusters occupy distinct, geometrically coherent regions of the $C_L-C_Y$ response plane.
Events at high lift with near-zero side force are captured by clusters concentrated in the
upper-right region of the plane, while the strongly asymmetric lateral-load events populate the
lower portion. 
The intermediate clusters span the curved central band, capturing transitional types between these
regimes. 
The \textbook{} centroids, marked by large symbols, sit near the geometric centers of each cluster. 
This is consistent with each \textbook{} event being representative of its assigned group rather
than an outlier.  
Taken together, the nine clusters suggest that the gust-load response space is structured into
physically distinct regimes that the \textbook{} has successfully identified.

The cluster membership per attitude is shown in \cref{fig:clusters}b and tabulated in
\cref{tab:cluster_membership}. 
Types 1, 9, 3 and 7 are exclusive to the two zero-sideslip attitudes ($\beta$ = $0^\circ$) and
occur at both angles of attack.
Types 2, 5, 6, 8, 9 are associated with non-zero sideslip conditions, with 2 and 5 being the only
types that only occur at a single attitude.
The zero sideslip angle $\beta=0$ is a strong qualitative differentiator of the response as it
implies $C_Y=0$, a property which is correctly detected by the objective summarization procedure.
This is consistent with the finding of \cref{sec:textbook_results} that attitude labeling does not
affect \textbook{} quality. 
Types 8 and 9 occur at four angles of attack and represent the most commonly occurring gust-load
types. 
The presence of cross-attitude clusters confirms that a subset of fundamental response types recurs
independently of the flight configuration.
The exemplar-based classification exposes structure of the response space that a parametric
criterion based on the attitude would obscure. 

\begin{table}[h]
\caption{
Event type occurrence across flight attitudes. 
A solid square indicates the cluster contributes more than 5\% of events at that attitude.
}
\label{tab:cluster_membership}
\small
\begin{tabular}{lcccccc}
\toprule
Type & \multicolumn{2}{c}{$\beta = 0^\circ$} & \multicolumn{2}{c}{$\beta = 15^\circ$} &
\multicolumn{2}{c}{$\beta = 30^\circ$} \\
 & $\alpha=20^\circ$ & $\alpha=30^\circ$ & $\alpha=20^\circ$ & $\alpha=30^\circ$ & $\alpha=20^\circ$
 & $\alpha=30^\circ$ \\
\midrule
1 & $\blacksquare$ & \qedsymbol & \qedsymbol & \qedsymbol & \qedsymbol & \qedsymbol \\
2 & \qedsymbol & \qedsymbol & \qedsymbol & $\blacksquare$ & \qedsymbol & \qedsymbol \\
3 & $\blacksquare$ & $\blacksquare$ & \qedsymbol & \qedsymbol & \qedsymbol & \qedsymbol \\
4 & \qedsymbol & \qedsymbol & $\blacksquare$ & $\blacksquare$ & \qedsymbol & $\blacksquare$ \\
5 & \qedsymbol & \qedsymbol & $\blacksquare$ & \qedsymbol & $\blacksquare$ & \qedsymbol \\
6 & \qedsymbol & \qedsymbol & $\blacksquare$ & $\blacksquare$ & $\blacksquare$ & $\blacksquare$ \\
7 & $\blacksquare$ & $\blacksquare$ & \qedsymbol & \qedsymbol & \qedsymbol & \qedsymbol \\
8 & \qedsymbol & \qedsymbol & $\blacksquare$ & $\blacksquare$ & $\blacksquare$ & $\blacksquare$ \\
9 & $\blacksquare$ & \qedsymbol & \qedsymbol & \qedsymbol & \qedsymbol & \qedsymbol \\
\botrule
\end{tabular}
\end{table}

\subsection{Characterization of \textbook{} events}
\label{sec:characterization}


The \textbook{} exemplars identified above organize the diversity of the wing's gust-load response
in terms of objective response types. A physically interesting question is investigating the nature
of the flow phenomena that generate a certain response type. This is especially true for explaining
how similar response types arise at different attitudes, which would be valuable physical insight
into the complexity of the wing's unsteady aerodynamics. 

As no direct measurements of the flow field was attempted, we resort to the closest available proxy,
i.e. the time histories of pressure and loading transients. 
As the position of the taps on the wing is fixed, the pressure time histories provide a sparse view
of the time-varying pressure field on the wing surface.
The force time histories (surface integrals of the pressure) subsume global pressure information. 

\begin{figure}[h]
\centering
\includegraphics[width=\linewidth]{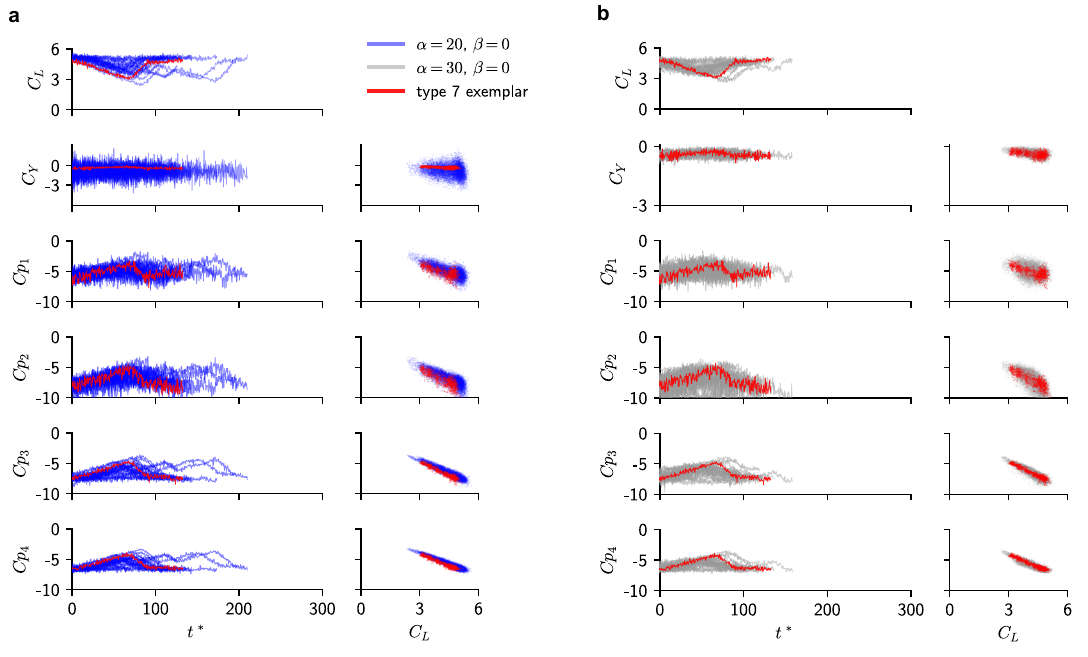}
\caption{\textbf{\textbook{} event signatures.}
Pressure-load histories for response type 7 and their signatures in input-response space.
Different colors indicate the original attitude each event was recorded at.
}
\label{fig:timeseries}
\end{figure}

As a study case, we consider responses of type 7 (highlighted in green in \cref{fig:clusters}),
which give rise to the simultaneously strongest $C_L-C_Y$ responses and belong to two different
attitudes, namely those where the sideslip angle $\beta=0$. 
The time histories of type-7 events is visually inspected in \cref{fig:timeseries} alongside the
type exemplar, which belongs to $\alpha=30$. 

We concentrate the analysis on two aspects: 
first, we recall from \cref{fig:clusters} that around $90\%$ of type 7 events occurs at $\alpha=30$; 
second, we notice from \cref{fig:timeseries} that the time histories of type-7 events share
attitude-wise qualitative commonalities. 
Specifically, type 7 load histories from $\alpha=30, \, \beta=0$ are qualitatively well represented
by the type exemplar, being generally single-peaked and with a median duration of $108 t^*$ and
median peak time at $53 t^*$. 
The other attitude $\alpha=20, \, \beta=0$ show generally longer median durations of $106 t^*$ and
predominantly multi-peaked or sawtooth-shaped histories. 
$49 t^*$
These facts hint to two different mechanisms leading to type-7 responses depending on the attack
angle. 

To this end, we consider existing PIV measurements of the flow around the same delta wing model
taken from a previous study \citep{Marzanek2019}, which were collected at an attitude of $\alpha=30,
\, \beta=0$ whilst the wing was subjected to a impulsive forward acceleration simulating an
encounter with a longitudinal gust.  
The PIVs exhibited leading-edge vortex (LEV) formation, followed by pinch-off which causes the flow
to  reattach and the lift force to revert to the baseline. 
These time-dependent features can be larger or smaller scale, depending on the attitude and the
magnitude of the acceleration, which results on average in similar instantaneous response. 

The observation that the higher angle of attack is responsible for most type-7 events agrees with
the physical intuition that higher mean lift must be more common; analogously, formation and
pinch-off of isolated LEVs is more likely at higher $\alpha$. 
On the other hand, at $\alpha=20$ only a minority of the encountered gusts have sufficient intensity
to produce similarly high baseline lift; flow detachment appears to persist for longer and produce a
series of weaker LEVs which form and pinch off intermittently.


A conclusive investigation of the fluid mechanics responsible for the observed response features
would necessitate repeating the experiment and perform flow-field measurements. 
However, these more sophisticated and costlier experiments would be focused on the exemplar cases
rather than repeated for many different combinations of attitude and fan forcing.  
This possibility highlights another practical advantage of the proposed approach. 


\section{Conclusions}
\label{sec:conclusion}


We investigate the pressure and load transient events induced on a non-slender delta wing by 3479
random  gust encounters across six angles of incidence, measured thanks to a partially automated
facility. 
The instantaneous pressure-load response is complex and not trivially separated by the incidence
angles of the wing with respect to the flow, motivating the search for an objective classification
criterion. 

We contribute with three points of methodological novelty.



First, building on the concept of \textbook{} \citep{Olivucci2026}, we are able to isolate of nine
essential gust-load response types by means of a purely data-driven, objective procedure.
The procedure leverages an accurate response model learned from the full database to distill the
database diversity into a few high-quality exemplars. 
This is the first application of such data-driven techniques to experimental aerodynamics that we
are aware of and represents a principled improvement on the previous iteration. 
We also notice that quality and number of the \textbook{} exemplars are insensitive to whether
attitude labeling is included or not into the selection process, 
hinting that the intuitive attitude-based classification does not provide any information on the
load response that is not detected by the data-based procedure. 

Second, we construct the desired objective classification criterion by labeling observations
according to their closest \textbook{} exemplar. 
We find that most fundamental gust-load types are not confined to one attitude, confirming that it
is not the most compressive organizing parameter. 
While a zero sideslip angle segregates load types into two non-overlapping groups, loads of the same
type occur at up to four different combinations of sideslip and angle of attack. 

Third, the manageable number of just a few response types is advantageous for mapping the unsteady
flow physics responsible for the aerodynamic response variety. 
A campaign of high-quality flow-field measurements could now be focused on the representative
cases rather than distributed across the experimental parameter range following a partial or
intuitive understanding of the response dependence. 
We examine the transient load histories of type 7, which reveal both common and exclusive
characteristics depending on the angle of attack. 
Existing flow-field data on the same wing model suggest the observed patterns are consistent with
the dynamics of leading-edge vortex formation and pinch off.   

In conclusion, this study provides a demonstration of a data-based framework for unfolding complex
aerodynamic phenomena which goes beyond sheer predictive power and emphasizes simplicity and
human-centric understanding. 
We believe this and related approaches have the potential to benefit unsteady aerodynamics, as well
as other areas of fluid mechanics where complex flow configurations are a primary concern.

\backmatter

\bmhead{Acknowledgements}

\section*{Declarations}

\textbf{Funding:}

\textbf{Conflict of interest:}
The authors declare no conflict of interest.

\textbf{Ethics approval:} 
Not applicable.

\textbf{Consent for publication:} 
Not applicable.

\textbf{Data availability:}
Upon reasonable request.

\textbf{Code availability:}
The code used for the data analysis will be available on GitHub upon publication of the manuscript.

\textbf{Author contribution:}
P.O.\ and K.S.\ contributed to the data analysis methodology. 
D.E.R.\ conceived the project and acquired funding. 
All authors contributed to the manuscript.

\bibliography{attitudes}

@Article{Gursul2005,
  author    = {Gursul, I. and Gordnier, R. and Visbal, M.},
  title     = {Unsteady aerodynamics of nonslender delta wings},
  doi       = {10.1016/j.paerosci.2005.09.002},
  issn      = {0376-0421},
  number    = {7},
  pages     = {515--557},
  volume    = {41},
  journal   = {Progress in Aerospace Sciences},
  month     = oct,
  publisher = {Elsevier BV},
  year      = {2005},
}

@Report{Moorhouse1982,
  author = {Moorhouse, David J. and Woodcock, Robert J.},
  title  = {Background Information and User Guide for MIL-F-8785C, "Military Specification - Flying Qualities of Piloted Airplanes". ADA119421},
  school = {Air Force Wright Aeronautical Labs},
  year   = {1982},
}

@InProceedings{Sener2018,
  author    = {Ozan Sener and Silvio Savarese},
  booktitle = {International Conference on Learning Representations},
  title     = {Active Learning for Convolutional Neural Networks: A Core-Set Approach},
  doi       = {10.48550/arXiv.1708.00489},
  year      = {2018},
}

@Article{Jones2022,
  author    = {Jones, Anya R. and Cetiner, Oksan and Smith, Marilyn J.},
  title     = {Physics and Modeling of Large Flow Disturbances: Discrete Gust Encounters for Modern Air Vehicles},
  doi       = {10.1146/annurev-fluid-031621-085520},
  issn      = {1545-4479},
  number    = {1},
  pages     = {469--493},
  volume    = {54},
  fjournal  = {Annual Review of Fluid Mechanics},
  journal   = {Annu. Rev. Fluid Mech.},
  month     = jan,
  publisher = {Annual Reviews},
  year      = {2022},
}

@Article{Jones2021,
  author    = {Jones, Anya R. and Cetiner, Oksan},
  title     = {Overview of Unsteady Aerodynamic Response of Rigid Wings in Gust Encounters},
  doi       = {10.2514/1.j059602},
  issn      = {1533-385X},
  number    = {2},
  pages     = {731--736},
  volume    = {59},
  fjournal  = {AIAA Journal},
  journal   = {AIAA J.},
  month     = feb,
  publisher = {American Institute of Aeronautics and Astronautics (AIAA)},
  year      = {2021},
}

@Article{Marzanek2019,
  author   = {Marzanek, M. F. and Rival, D. E.},
  title    = {Separation mechanics of non-slender delta wings during streamwise gusts},
  doi      = {10.1016/j.jfluidstructs.2019.07.001},
  issn     = {0889-9746},
  pages    = {286-296},
  volume   = {90},
  fjournal = {Journal of Fluids and Structures},
  journal  = {J. Fluids Struct.},
  year     = {2019},
}

@InProceedings{Paul2021,
  author    = {Paul, Mansheej and Ganguli, Surya and Dziugaite, Gintare Karolina},
  booktitle = {Advances in Neural Information Processing Systems},
  title     = {Deep Learning on a Data Diet: Finding Important Examples Early in Training},
  doi       = {10.48550/arXiv.2107.07075},
  editor    = {M. Ranzato and A. Beygelzimer and Y. Dauphin and P.S. Liang and J. Wortman Vaughan},
  pages     = {20596--20607},
  publisher = {Curran Associates, Inc.},
  volume    = {34},
  year      = {2021},
}

@Article{Burelle2020,
  author   = {Burelle, L. A. and others},
  title    = {{Exploring the signature of distributed pressure measurements on non-slender delta wings during axial and vertical gusts}},
  doi      = {10.1063/5.0025860},
  issn     = {1070-6631},
  number   = {11},
  pages    = {115110},
  volume   = {32},
  fjournal = {Physics of Fluids},
  journal  = {Phys. Fluids},
  month    = nov,
  year     = {2020},
}

@Article{Fuller1995,
  author    = {Fuller, J. R.},
  title     = {Evolution of airplane gust loads design requirements},
  doi       = {10.2514/3.46709},
  issn      = {1533-3868},
  number    = {2},
  pages     = {235--246},
  volume    = {32},
  fjournal  = {Journal of Aircraft},
  journal   = {J. Aircraft},
  month     = mar,
  publisher = {American Institute of Aeronautics and Astronautics (AIAA)},
  year      = {1995},
}

@Article{Floreano2015,
  author    = {Floreano, Dario and Wood, Robert J.},
  title     = {Science, technology and the future of small autonomous drones},
  doi       = {10.1038/nature14542},
  issn      = {1476-4687},
  number    = {7553},
  pages     = {460--466},
  volume    = {521},
  journal   = {Nature},
  month     = may,
  publisher = {Springer Science and Business Media LLC},
  year      = {2015},
}

@Article{Fukami2023,
  author    = {Fukami, Kai and Taira, Kunihiko},
  title     = {Grasping extreme aerodynamics on a low-dimensional manifold},
  doi       = {10.1038/s41467-023-42213-6},
  issn      = {2041-1723},
  number    = {1},
  volume    = {14},
  fjournal  = {Nature Communications},
  journal   = {Nat. Commun.},
  month     = oct,
  publisher = {Springer Science and Business Media LLC},
  year      = {2023},
}

@Article{Chen2023,
  author   = {Chen, D. and Kaiser, F. and Hu, J. C. and Rival, D. E. and Fukami, K. and Taira, K.},
  title    = {Sparse pressure-based machine learning approach for aerodynamic loads estimation during gust encounters},
  doi      = {10.2514/1.J063263},
  fjournal = {AIAA Journal},
  journal  = {AIAA J.},
  year     = {2023},
}

@Article{Iacobello2022,
  author    = {Iacobello, Giovanni and Kaiser, Frieder and Rival, David E.},
  title     = {Load estimation in unsteady flows from sparse pressure measurements: Application of transition networks to experimental data},
  doi       = {10.1063/5.0076731},
  issn      = {1089-7666},
  number    = {2},
  volume    = {34},
  fjournal  = {Physics of Fluids},
  journal   = {Phys. Fluids},
  month     = feb,
  publisher = {AIP Publishing},
  year      = {2022},
}

@Article{Olivucci2026,
  author        = {Olivucci, Paolo and Rival, David E.},
  title         = {The search for the gust-wing interaction "textbook"},
  doi           = {10.48550/ARXIV.2602.10968},
  eprint        = {2602.10968},
  archiveprefix = {arXiv},
  copyright     = {arXiv.org perpetual, non-exclusive license},
  month         = feb,
  primaryclass  = {physics.flu-dyn},
  publisher     = {arXiv},
  year          = {2026},
}

@Article{Leishman1996,
  author    = {Leishman, J. Gordon},
  title     = {Subsonic unsteady aerodynamics caused by gusts using the indicial method},
  doi       = {10.2514/3.47029},
  issn      = {1533-3868},
  number    = {5},
  pages     = {869--879},
  volume    = {33},
  journal   = {Journal of Aircraft},
  month     = Sept,
  publisher = {American Institute of Aeronautics and Astronautics (AIAA)},
  year      = {1996},
}

@Article{Gunasekar2023,
  author        = {Gunasekar, S. and others},
  title         = {Textbooks Are All You Need},
  doi           = {10.48550/arxiv.2306.11644},
  archiveprefix = {arXiv},
  journal       = {arXiv:2306.11644v2},
  year          = {2023},
}

@Article{Kaiser2024,
  author    = {Kaiser, Frieder and Iacobello, Giovanni and Rival, David E.},
  title     = {Cluster-based Bayesian approach for noisy and sparse data: application to flow-state estimation},
  doi       = {10.1098/rspa.2023.0608},
  issn      = {1471-2946},
  number    = {2291},
  volume    = {480},
  journal   = {Proceedings of the Royal Society A: Mathematical, Physical and Engineering Sciences},
  month     = jun,
  publisher = {The Royal Society},
  year      = {2024},
}

@InProceedings{Koh17,
  author    = {Pang Wei Koh and Percy Liang},
  booktitle = {Proceedings of the 34th International Conference on Machine Learning},
  title     = {Understanding Black-box Predictions via Influence Functions},
  doi       = {10.48550/arXiv.1703.04730},
  editor    = {Precup, Doina and Teh, Yee Whye},
  pages     = {1885--1894},
  publisher = {PMLR},
  series    = {Proceedings of Machine Learning Research},
  volume    = {70},
  year      = {2017},
}

@Article{Tansel1983,
  author       = {Tansel, Barbaros C. and Francis, Richard L. and Lowe, Timothy J.},
  date         = {1983-04},
  journaltitle = {Management Science},
  title        = {State of the Art—Location on Networks: A Survey. Part I: The $p$-Center and $p$-Median Problems},
  doi          = {10.1287/mnsc.29.4.482},
  issn         = {1526-5501},
  number       = {4},
  pages        = {482--497},
  volume       = {29},
  publisher    = {Institute for Operations Research and the Management Sciences (INFORMS)},
}

\appendix
\section{Appendix}\label{sec:appendix}

\subsection{Trial segmentation}\label{sec:segmentation} 
The procedure to segment continuous trials into discrete events is as follows.
The mode of $C_L$ across each trial is computed as a proxy for the baseline flow state, and
candidate event boundaries are identified at points where $C_L$ reverses to this baseline. 
Each candidate segment is subsequently assessed against four quality thresholds on its maximum and
minimum duration, noise-to-signal ratio and monotonicity, and discarded if it fails to satisfy any
of them. 

\begin{figure}[h]
\centering
\includegraphics[width=0.9\linewidth]{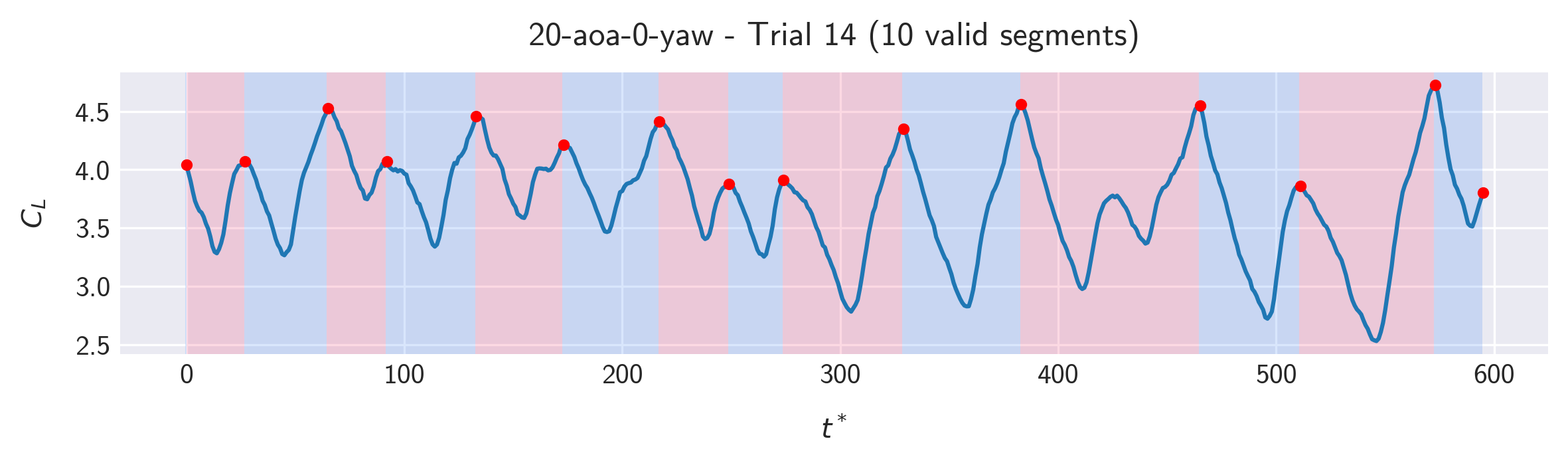}
\caption{Segmentation of a representative one-minute trial of $C_L(t)$ into individual gust events. 
Valid segments are shown against a colored background; red dots mark event boundaries.  
}  
\label{fig:segmentation}
\end{figure}


\end{document}